\documentclass{article}

 \usepackage[preprint]{neurips_2026}

\usepackage[utf8]{inputenc} 
\usepackage[T1]{fontenc}    
\usepackage{hyperref}       
\usepackage{url}            
\usepackage{booktabs}       
\usepackage{amsfonts}       
\usepackage{nicefrac}       
\usepackage{microtype}      
\usepackage{xcolor}         
\usepackage{multirow}
\usepackage{amsmath}
\usepackage{graphicx}
\usepackage{cleveref}
\title{Beyond Facial Consistency: Personalized Person Image Generation with Holistic Identity Preservation}

\author{%
  Yuxuan Xiao 
  \And
  Shanshan Zhang
  \And
  Jian Yang
  \And
  Shengcai Liao
}

\begin{document}

\maketitle

\begin{abstract}
    Personalized person image generation requires preserving subject identity across both local facial details and broader appearance cues. Existing methods typically emphasize only one level of identity information, leading to an inherent trade-off between facial fidelity and overall appearance consistency. To address this, we first propose a simple dual-branch baseline that unifies global appearance control and local facial control within a shared generation framework. 
    This simple combination of different branches yields promising results, but suffers from instability in practice due to uncoordinated branch contributions.
    To this end, we propose Dynamic Balancing Scaling (DBS), a fine-tuning strategy for improving face and appearance identity coordination. DBS consists of two components: adaptive temporal gating, which dynamically modulates branch contributions along the denoising trajectory, and region-aware optimization, which improves the coordination of facial, appearance, and global supervision. Together, these designs alleviate persistent face-branch over-dominance and encourage more effective appearance-aware guidance.
    We also introduce Pexels-100, a benchmark for evaluating holistic identity consistency in personalized person generation. Experiments show that DBS achieves a better trade-off between facial fidelity and appearance consistency than existing open-source baselines, while providing a controllable basic framework for holistic identity modeling.
\end{abstract}

\section{Introduction}
\begin{figure}[t]
\begin{center}
  \includegraphics[width=0.99\linewidth]{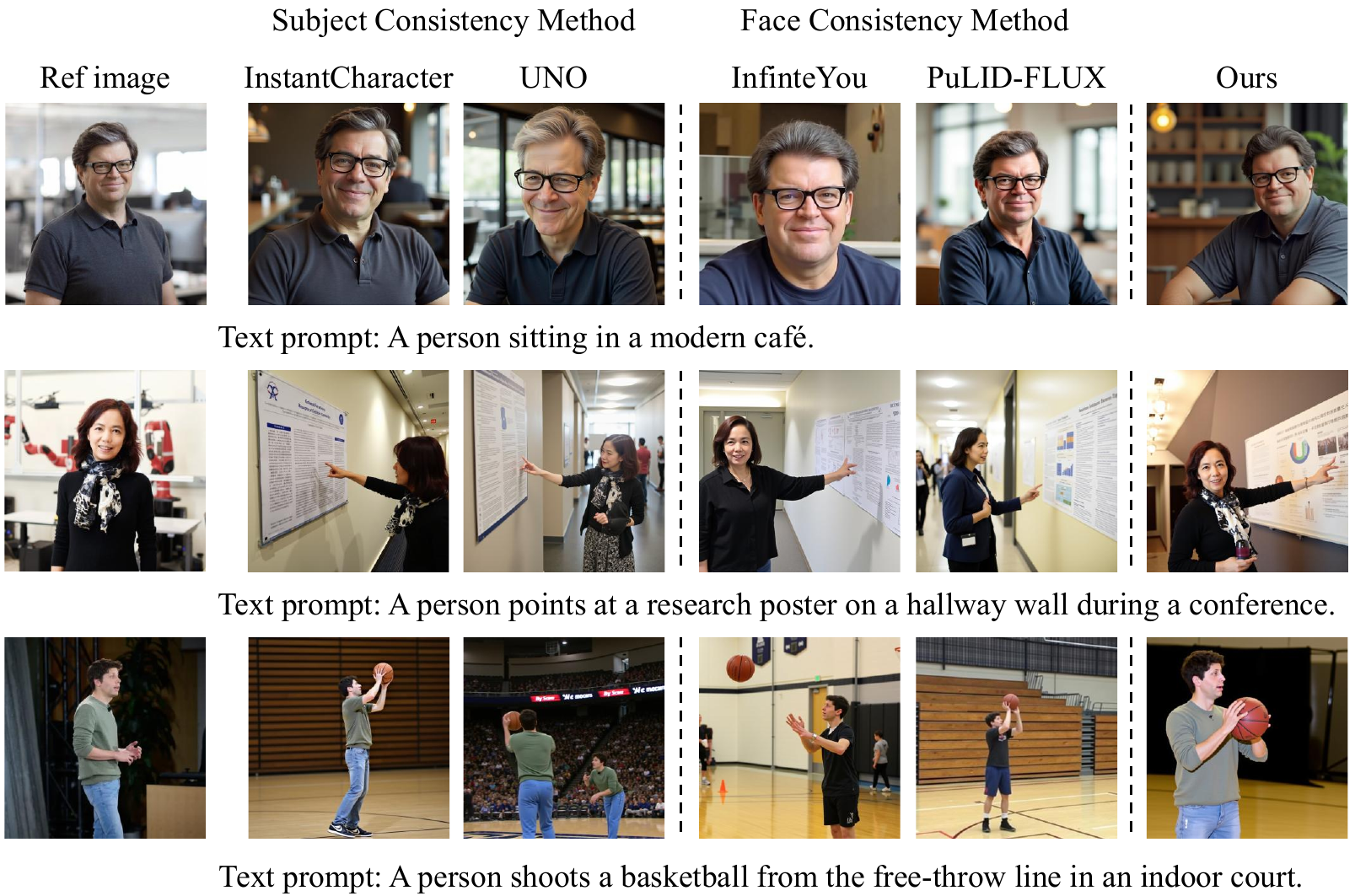}
\end{center}
  \caption{Comparison of our method with subject consistency and face consistency methods. Please zoom in for better visualization. Best viewed in color.}
  \label{fig:intro}
\end{figure}

Personalized person image generation aims to generate images that closely match the user's identity features based on a user-provided reference image and text prompts. This technology has significant applications in film and television production, commercial advertising, artistic creation, and digital human generation.
Existing methods typically focus only on a portion of the identity's attributes. As shown in~\cref{fig:intro}, one type of method emphasizes subject consistency, such as character or object; another type focuses on preserving local identity, especially the fidelity of facial structure and details.
However, in real-world applications, single-dimensional identity consistency is insufficient to meet complex needs. For example, in continuous content generation, digital human creation, and personalized business applications, models need to maintain a stable overall identity expression across different scenarios. This involves not only facial features but also multi-layered semantic information such as clothing style, body shape, and overall visual presentation.
Therefore, it is necessary to research a unified generation framework that can simultaneously constrain facial and full-body appearance from the perspective of holistic identity modeling.

\begin{figure}[t]
\begin{center}
  \includegraphics[width=0.99\linewidth]{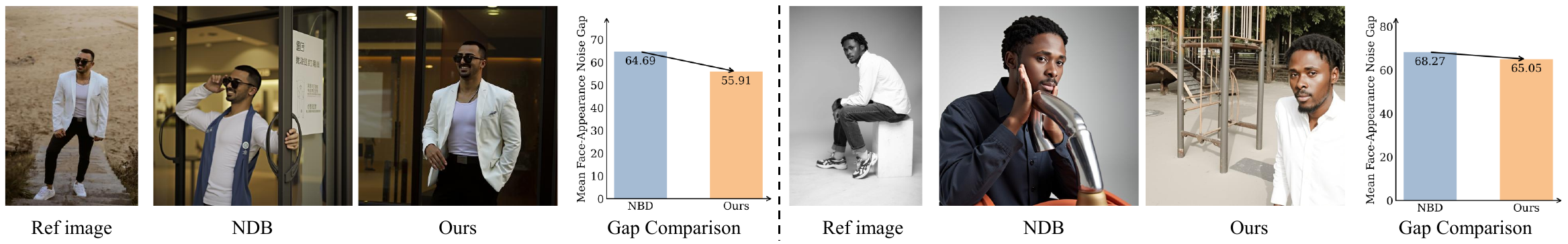}
\end{center}
  \caption{Our method improves appearance consistency while maintaining facial stability by coordinating different branches. Please zoom in for better visualization. Best viewed in color.}
  \label{fig:gap}
\end{figure}

To address this challenge, we propose modeling the person's appearance and local facial features separately and introduce a naive dual-branch (NDB) method, which integrate a subject (appearance) consistency method~\cite{tao2025instantcharacter} and a face consistency method~\cite{jiang2025infiniteyou}. 
Both methods are built upon the FLUX.1-dev~\cite{flux2024} backbone.
This simple combination of reference conditioning yields promising results, but suffers from instability in practice. As illustrated in~\cref{fig:gap}, the model tends to preserve facial identity well, while appearance consistency is comparatively less stable.
In order to investigate this phenomenon, we independently remove the face branch and the appearance branch, and for each setting we measure the magnitude of the resulting change in noise prediction relative to the full model during denoising. We then compare these two perturbation magnitudes and report their average difference over all denoising steps.
As observed in~\cref{fig:gap} (Gap Comparison), the average difference remains substantially above zero, meaning that removing the face branch perturbs the model prediction much more strongly than removing the appearance branch. This reveals a heavier reliance on facial guidance and relatively weaker appearance constraints in the generation process.
This imbalance arises from the heterogeneous representation spaces and injection mechanisms of the two branches, which lead to different interaction strengths with the DiT backbone. Explicit coordination of branch contributions is therefore necessary to improve holistic identity consistency while preserving stable facial identity.

To this end, we build upon the dual-branch baseline and further propose a Dynamic Balancing Scaling (DBS) fine-tuning strategy to better coordinate identity information between face and appearance throughout the generation process. Specifically, DBS consists of two complementary components:
(1) Adaptive Temporal Gating: dynamically modulating the contribution strength of different identity branches over the denoising trajectory. By introducing a time-aware gating mechanism, the influences of the Appearance branch and the Face branch are adaptively adjusted at different timesteps, which improves generation stability and promotes more consistent identity preservation across global appearance and local facial details.
(2) Region-Aware Optimization: explicitly regularizing branch interactions between face and appearance. We design a region-aware optimization objective to alleviate the over-dominance of face-related signals and encourage more effective contribution from appearance-related cues, thereby reducing branch imbalance and mitigating mutual interference between local identity cues and broader appearance information.
To fairly and accurately evaluate the holistic consistency modeling of different methods, we constructed the Pexels-100 test set, the first test benchmark Pexels-100 for maintaining holistic consistency in personalized person image generation. Qualitative and quantitative experimental results show that the proposed method outperforms existing open-source methods in terms of holistic identity consistency and reaches or exceeds the performance of some closed-source models on multiple evaluation metrics.

Our contributions are summarized as follows:
\begin{itemize}
\item To maintain the holistic identity consistency of the reference person in personalized person image generation, we propose explicitly modeling the person's appearance and local facial features separately and introduce a naive dual-branch (NDB) method.
\item We propose a Dynamic Balanced Scaling (DBS) fine-tuning strategy, which alleviates the problem of insufficient holistic consistency maintenance under strong signal dominance in NDB through adaptive temporal gating mechanism and region-aware optimization.
\item We construct the first benchmark, Pexels-100, for evaluating holistic consistency in personalized person image generation. Extensive evaluation on this benchmark demonstrates that our approach achieves more stable and consistent identity preservation.
\end{itemize}

\section{Related Work}
\subsection{Subject Consistency}
Subject consistency aims to maintain the overall visual consistency of a reference subject under varying conditions during the generation process.
Existing methods in this direction can be broadly categorized into two groups. One group focuses on character or human appearance references, while the other emphasizes more general subject references or conditional control.
Representative character consistency methods include InstantCharacter~\cite{tao2025instantcharacter}, which targets open-domain character customization and achieves high-fidelity consistency across diverse appearances, poses, and styles. Visual Persona~\cite{nam2025visual} further extends this line of work to full-body human customization, emphasizing the generation of fully personalized results from a single outdoor image, and leveraging region-level modeling to improve appearance transfer.
In contrast, more general subject control methods include EasyControl~\cite{zhang2025easycontrol}, which provides efficient and flexible unified conditional control for diffusion transformers; OminiControl~\cite{tan2025ominicontrol}, which achieves unified modeling across multiple control tasks with minimal parameter overhead; UNO~\cite{wu2025less}, which improves consistency and controllability in both single- and multi-subject settings via multi-image conditioning; and FLUX.1 Kontext~\cite{labs2025kontext}, which supports diverse in-context tasks such as character reference, style reference, and local/global editing within a unified framework.

Despite these advances, most existing methods focus on consistency at a single level or on general subject control, and do not explicitly address holistic identity consistency. In particular, they lack hierarchical modeling that jointly captures global appearance and fine-grained facial details. As a result, these approaches remain limited in personalized character generation scenarios that require simultaneous preservation of character attributes, overall appearance, and facial identity.

\subsection{Face Consistency}
Compared to subject consistency, which focuses on the overall appearance of the subject, face consistency methods primarily use a reference face as an identity condition, preserving the facial identity features of the person during the generation process. This direction is relatively mature and has gradually formed a technological evolution path from general image cues to dedicated face identity modeling.
In early work, IP-Adapter~\cite{ye2023ip} proposed a lightweight adaptation method that incorporates image cues into diffusion models, providing a foundational capability for subsequent identity preservation tasks by decoupling text and image conditions. Subsequently, a series of methods began to focus on face identity modeling. For example, InstantID~\cite{wang2024instantid} achieves zero-shot identity-preserving generation using a single face image, enhancing person consistency without additional training; ConsistentID~\cite{huang26consistentid} further improves identity fidelity in portrait generation through fine-grained facial feature modeling.
Building on this foundation, methods have gradually introduced stronger identity representation capabilities. PhotoMaker~\cite{li2024photomaker} constructs a more stable identity representation by encoding multiple reference faces, thereby improving identity consistency in text-driven generation. PuLID~\cite{guo2024pulid} proposes a tuning-free identity customization framework based on contrastive alignment, enabling high-fidelity identity preservation while maintaining the original generation behavior. InfiniteYou~\cite{jiang2025infiniteyou} further emphasizes maintaining identity under free-text control, achieving more flexible identity-preserving generation. Additionally, FlashFace~\cite{zhang2024flashface} also achieves high-fidelity image generation using a small number of facial references.

These methods have made significant progress in facial identity preservation, but they mainly focus on the facial region and rely primarily on facial reference images. As a result, they are more effective at maintaining facial identity, but remain limited in modeling overall appearance, clothing style, and holistic visual expression, making it difficult to directly meet the requirements of personalized person generation with holistic identity consistency.

\begin{figure}[t]
\begin{center}
  \includegraphics[width=0.99\linewidth]{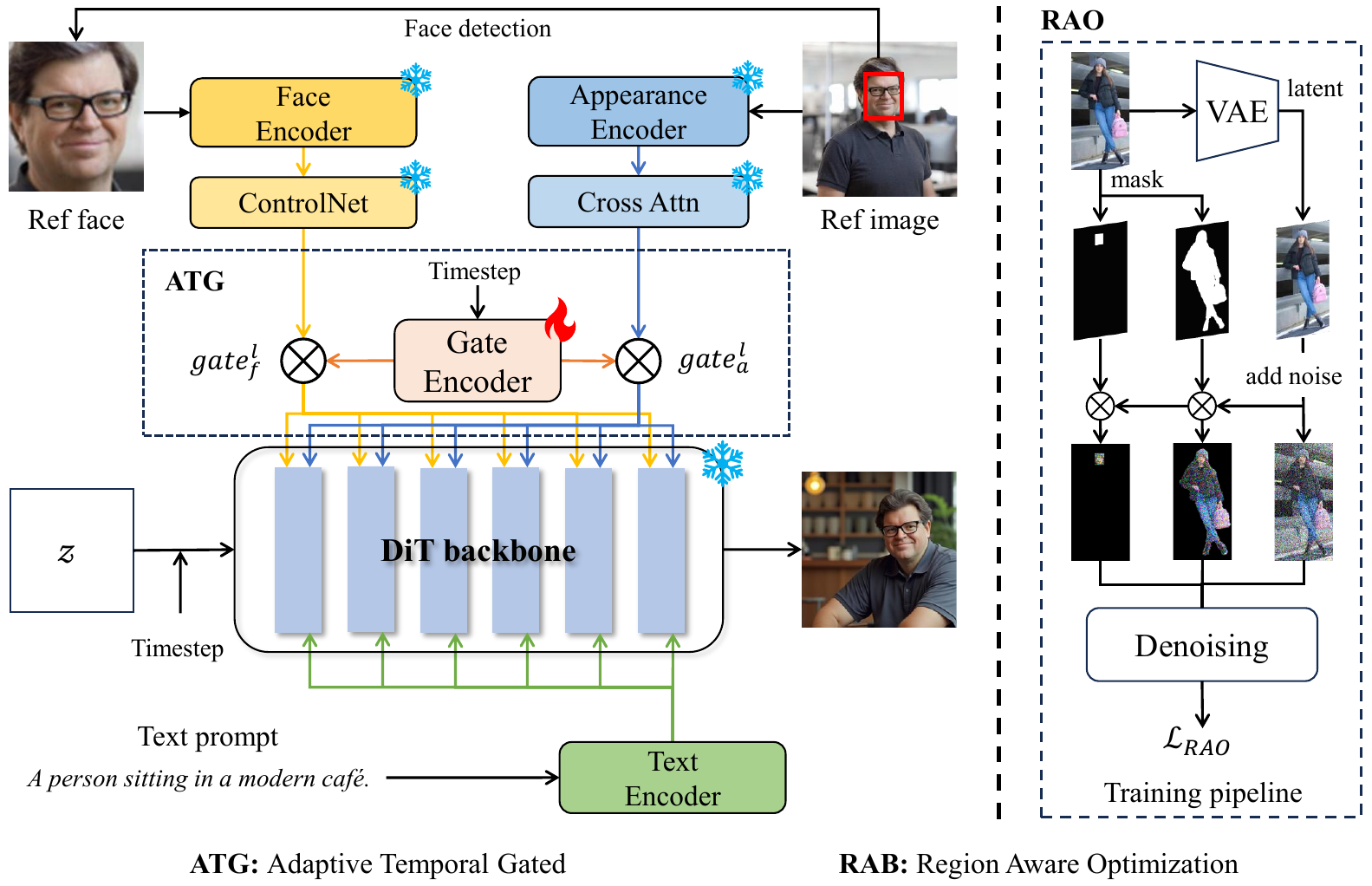}
\end{center}
  \caption{Overview of our framework. Built upon a naive dual-branch architecture, DBS introduces Adaptive Temporal Gating (ATG) to adaptively modulate branch contributions during generation, together with Region Aware Optimization (RAO) to improve cross-granularity coordination, alleviate the over-dominance of face-related signals, and encourage more effective appearance-aware guidance.}
  \label{fig:overview}
\end{figure}

\section{Methodology}
\subsection{Overview}
We aim to achieve holistic identity consistency in the context of personalized person image generation. Given a reference identity image $I_{ref}$ and a target text prompt $P$ (note that our prompts do not contain any direct descriptions of the person, such as appearance or gender, to avoid providing additional information beyond the reference image.), the objective is to synthesize a high-fidelity image $I_{gen}$ that simultaneously preserves global appearance (e.g., body shape and clothes) and local facial identity (e.g., fine-grained facial features). 
As shown in~\cref{fig:overview}, we first establish a naive dual-branch architecture based on~\cite{tao2025instantcharacter} and~\cite{jiang2025infiniteyou}. However, due to the inherent heterogeneity between branches, this naive combination introduces competing signals between face and appearance, which leads to unstable generation. To address this issue, we propose a Dynamic Balancing Scaling (DBS) strategy. Specifically, DBS first introduces a temporal gating mechanism to adaptively modulate the contributions of different branches during generation. It then incorporates a region-aware optimization mechanism to improve cross-granularity coordination, alleviating the over-dominance of face-related signals while encouraging more effective contribution from broader appearance cues.

\subsection{Naive Dual-Branch}
To establish a baseline for holistic identity preservation, we extend the FLUX.1-dev~\cite{flux2024} rectified flow transformer with two complementary identity-conditioning pathways. Rather than treating identity as a single homogeneous signal, our baseline explicitly separates it into two levels of granularity: a global appearance pathway for coarse identity attributes and a local facial pathway for fine-grained facial cues. Both pathways are adapted to a shared generative backbone and jointly participate in the denoising process, forming a naive yet effective starting point for studying holistic identity preservation.

\textbf{Appearance Branch}~\cite{tao2025instantcharacter}. The appearance branch captures global identity cues, including hairstyle, body proportion, and clothing style, which together characterize the subject's overall visual appearance. Specifically, DINOv2~\cite{oquab2024dinov2} and SigLIP~\cite{zhai2023sigmoid} features are fused via channel-wise concatenation to obtain a more comprehensive representation for open-domain character modeling. The fused representation is then projected into the denoising space through a timestep-aware Q-Former~\cite{li2023blip} projector, and injected into the backbone via cross-attention, thereby providing global guidance for overall structure and appearance generation.

\textbf{Face Branch}~\cite{jiang2025infiniteyou}. To complement the appearance pathway, the face branch focuses on fine-grained facial geometry and identity-sensitive texture details. Facial identity embeddings are first extracted by a pretrained face encoder and projected into a compatible feature space. These embeddings are then injected into the DiT backbone through a ControlNet~\cite{zhang2023adding}-style residual pathway, enabling stable facial identity guidance throughout the denoising process.

\subsection{Adaptive Temporal Gating}
As discussed before, although the naive dual-branch approach yields promising results, it remains unstable because branch heterogeneity gives rise to inconsistent contribution strengths and uncoordinated interactions during denoising. 
To alleviate this issue, we propose a timestep-aware adaptive gating mechanism, which improve the coordination between branches over the denoising trajectory and reduce persistent over-reliance on a single branch.
In particular, we use the face branch as a stable reference for optimization, and progressively modulate the appearance branch relative to this reference, thereby stabilizing early training while preserving sufficient flexibility for branch-specific adaptation.

Formally, the diffusion timestep $t$ is first projected into a high-dimensional temporal embedding via sinusoidal encoding and then processed by a lightweight mapping network to obtain a latent representation $\mathbf{h}_t$:
\begin{equation}
\mathbf{h}_t = \phi(\mathrm{Embed}(t)),
\end{equation}
where $\mathrm{Embed}(\cdot)$ denotes the timestep encoding function and $\phi(\cdot)$ denotes the mapping network.

For the face branch, we define a bounded residual gate centered around unity:
\begin{equation}
gate^{(l)}_{f}(t) = 
1 + \delta \tanh(\mathbf{W}^{(l)}_{f} \mathbf{h}_t + \mathbf{b}^{(l)}_{f}),
\end{equation}
where $\delta$ controls the modulation range and $l$ indexes the layer. This formulation keeps the face branch close to its original operating regime, allowing it to provide stable identity guidance while still supporting moderate temporal adaptation.

For the appearance branch, instead of predicting a fully independent gate from scratch, we parameterize its contribution relative to the face branch:
\begin{equation}
gate^{(l)}_{a}(t; \tau) =
\frac{1}{L} \sum_{l=1}^{L} gate_{f}^{(l)}(t)
\cdot
\left(1 + \alpha(\tau)\gamma
\tanh(\mathbf{W}^{(l)}_{a}\mathbf{h}_t + \mathbf{b}^{(l)}_{a})\right),
\end{equation}
where $\gamma$ controls the deviation range, $\alpha(\tau)\in[0,1]$ is a progressive relaxation coefficient that increases with training progress $\tau$, and $L$ is the total number of layers. Here, the face branch serves as a reference anchor rather than a permanently dominant signal. The appearance branch is initialized around this reference to avoid unstable optimization caused by unconstrained branch competition, and is then gradually relaxed through $\alpha(\tau)$ to acquire branch-specific flexibility. In this way, the model can progressively strengthen appearance-aware guidance without introducing severe interference in early training.

In practice, the temporal gates are applied to the two branches separately. For the face branch, $gate_f(t)$ modulates the conditional residuals $\mathbf{F}(t)$ produced by ControlNet:
\begin{equation}
\mathbf{H}^{(l)} \leftarrow \mathbf{H}^{(l)} + gate^{(l)}_{f}(t)\,\mathbf{F}^{(l)}(t).
\end{equation}
For the appearance branch, $gate_a(t;\tau)$ scales the appearance-aware attention residuals $\mathbf{A}_a(t)$ before fusing them into the backbone:
\begin{equation}
\mathbf{H}^{(l)} \leftarrow \mathbf{H}^{(l)} + s_a \cdot gate^{(l)}_{a}(t;\tau)\,\mathbf{A}^{(l)}_{a}(t).
\end{equation}

Through this anchor-guided temporal modulation, the model improves the coordination of multi-granularity identity cues across diffusion stages, stabilizes optimization under heterogeneous conditioning, and alleviates persistent face-branch over-dominance while preserving effective appearance-aware guidance.

\subsection{Region Aware Optimization}
Identity-related facial regions typically occupy only a small portion of the image, while broader appearance cues are distributed across larger body-related regions. While adaptive temporal gating modulates the relative strength of different identity branches  the denoising trajectory, it cannot inherently perceive the spatial distribution of the portrait image. In this case of regional imbalance, the standard global mean squared error (MSE) objective function treats all pixels uniformly, providing spatially uniform supervision, which limits the gating function's ability to fully optimize identity cues in specific regions.

To this end, we introduce a region-aware optimization strategy that explicitly incorporates spatially structured supervision at different identity information. 
Specifically, we first construct a hierarchical masking system using pixel-level human segmentation and face detection to obtain a person region mask $M_a$ and a facial region mask $M_f$, where $M_f \subseteq M_a$. Based on these masks, the denoising error map $E$ is decomposed into three regional objectives:
\begin{equation}
\mathcal{L}_{f} = \mathrm{Mean}(M_f \odot E), \quad
\mathcal{L}_{a} = \mathrm{Mean}(M_a \odot E), \quad
\mathcal{L}_{g} = \mathrm{Mean}(E),
\end{equation}
where $\odot$ denotes the Hadamard product. Here, $\mathcal{L}_f$ emphasizes facial identity preservation, $\mathcal{L}_a$ supervises broader appearance-related regions, and $\mathcal{L}_g$ maintains global image quality.
Directly combining the three loss functions can be suboptimal. The facial objective is defined over a small yet highly sensitive identity region, resulting in concentrated supervision, whereas the appearance and global objectives provide broader but relatively smoother guidance. Without explicit coordination, such heterogeneous supervision makes it difficult for the model to maintain stable facial identity cues while effectively optimizing appearance consistency.

To coordinate these region-specific objectives, we further introduce a reference-guided Pareto-style loss aggregation approach.
Concretely, the face and appearance objectives are treated as protected identity-related terms: they are explicitly activated when their performance falls below a running reference level, while the global objective serves only as a persistent background objective for maintaining overall image quality. 
Given the regional losses $L_f^{(k)}$, $L_a^{(k)}$, and $L_g^{(k)}$ at training step $k$, we maintain an exponential moving average (EMA) reference for each objective:
\begin{equation}
\bar{L}_m^{(k+1)} = (1 - \mu)\bar{L}_m^{(k)} + \mu L_m^{(k)}, \quad m \in \{f, a, g\}.
\end{equation}
We then normalize each loss as
\begin{equation}
r_m^{(k)} = \frac{L_m^{(k)}}{\bar{L}_m^{(k)} + \epsilon}.
\end{equation}
To reflect their different functional roles, we define
\begin{equation}
v_m^{(k)} =
\begin{cases}
\max(r_m^{(k)} - 1, 0), & m \in \{f, a\}, \\
r_m^{(k)}, & m = g,
\end{cases}
\end{equation}
so that the regional identity losses become active only when they underperform relative to their reference levels, whereas the global objective remains continuously optimized. The final objective is
\begin{equation}
\mathcal{L}_{RAO}^{(k)} =
\eta \log \sum_{m \in \{f, a, g\}}
\exp \left( \frac{v_m^{(k)}}{\eta} \right),
\end{equation}
where $\eta$ is a temperature parameter and is set to $0.1$.
This design continuously optimizes global coherence while explicitly protecting region-specific identity supervision, leading to more stable coordination among facial, appearance, and global objectives and alleviating optimization trajectories dominated by a single branch.

\section{Experiment}
\subsection{Implementation details}
We use FLUX.1-dev~\cite{flux2024} as the denoising backbone, and derive the appearance and face branches from~\cite{tao2025instantcharacter} and~\cite{jiang2025infiniteyou}, respectively. All experiments are conducted on a single NVIDIA A100 GPU with 80GB VRAM using the \texttt{accelerate} framework. The model is optimized with 8-bit AdamW~\cite{loshchilovdecoupled}, using $\beta_1=0.9$, $\beta_2=0.999$, and weight decay $0.01$. The gate parameters are trained with an initial learning rate of $5\times10^{-5}$. Training images are resized with aspect ratio preserved, using a maximum side length of $512$. The instantaneous batch size is set to $1$, and gradient accumulation over 2 steps is used to obtain an effective batch size of $2$. The model is trained for $50K$ iterations.

\subsection{Datasets and Evaluation}
\textbf{Datasets}. We use the real-world dataset PPR10K~\cite{liang2021ppr10k} as the training data. After filtering for single-face images and enforcing a face similarity threshold greater than 0.75, we constructed a training set of 10,444 pairs of images containing 850 identities.
We further annotated these image with textual descriptions using Qwen3-VL~\cite{bai2025qwen3}.
It's worth noting that, to avoid textual interference and leakage of reference information, the annotation process deliberately avoids including identity-related attributes such as appearance or gender. Instead, the descriptions focus solely on actions and scene context. For example, a typical annotation takes the form ``\textit{a person rests on a stone with a rose in hand, surrounded by flowers in a wooded area.}''

\textbf{Evaluation}.
Since existing identity evaluation protocols are primarily based on facial consistency, we construct a holistic identity consistency benchmark, Pexels-100, from full-body human images collected from the Pexels\footnote{\url{https://www.pexels.com/}}. We used ChatGPT~\cite{chatgpt2024} to generate 100 text prompts and paired them with these images.
To evaluate holistic identity consistency, we adopt face similarity, person re-identification (ReID) score, and DINOv2 score~\cite{oquab2024dinov2} as quantitative metrics. Face similarity is computed using ArcFace model~\cite{deng2019arcface}, while ReID score is measured using TransMatcher~\cite{liao2021transmatcher}, a cross-domain ReID model. In addition, CLIP-T~\cite{hessel2021clipscore} is introduced as a complementary measure for text-image alignment.

\begin{table}[t]
\centering
\caption{Quantitative comparison. Compared to non-commercial open-source models, our method achieves the best holistic identity consistency.}
\begin{tabular}{l lcccc}
\toprule
\textbf{Type} & \textbf{Method} & \textbf{Face} & \textbf{ReID} & \textbf{DINOv2} & \textbf{CLIP-T} \\
\midrule
\multirow{9}{*}{Open-source}
& EasyControl~\cite{zhang2025easycontrol}             & 0.0659 & 0.8967  & 0.1710 & 0.2196 \\
& OminiControl~\cite{tan2025ominicontrol}             & 0.0893 & 0.8999 & 0.1795 & \textbf{0.2360} \\
& UNO~\cite{wu2025less}                               & 0.1068 & 0.9105 & 0.2561 & 0.2212 \\
& Kontext~\cite{labs2025kontext}                      & 0.4910 & 0.9161 & 0.2964 & 0.2150 \\
& InstantCharacter~\cite{tao2025instantcharacter}     & 0.1514 & 0.9118 & 0.2300 & 0.2119 \\
& InfiniteYou~\cite{jiang2025infiniteyou}             & 0.7522 & 0.8852 & 0.2094 & 0.1678 \\
& Visual Persona~\cite{nam2025visual}                 & 0.2242 & 0.9102 & 0.2242 & 0.2357 \\
& Qwen-Image~\cite{wu2025qwen}                        & 0.4640 & \textbf{0.9230} & 0.2486 & 0.2279\\
& JoyAI-Image~\cite{joyai_image}                      & 0.3130 & \textbf{0.9230} & 0.2538 & 0.2219\\
\midrule
\multirow{1}{*}{Upcoming}
& Ours                                                & \textbf{0.7875} & 0.9122 & \textbf{0.3202} & 0.2008 \\
\midrule
\multirow{4}{*}{Closed-source} 
& Nano banana~\cite{nano_banana_2025}                 & 0.4948 & 0.9251 & 0.2316 & 0.2316\\
& Nano banana2~\cite{nano_banana2_2026}               & 0.5933 & 0.9187 & 0.2528 & 0.2260\\
& GPT-Image-1.5~\cite{gpt_image_1_5}                  & 0.3280 & 0.9213 & 0.3305 & 0.2258 \\
& GPT-Image-2~\cite{gpt_image_2}                      & 0.8519 & 0.9291 & 0.3278 & 0.2249  \\
\bottomrule
\end{tabular}
\label{tab:quantitative_comparison}
\end{table}

\subsection{Main Results}
\textbf{Quantitative comparison.}
We compare our approach with representative subject-consistency (Note that EasyControl~\cite{zhang2025easycontrol} is compared here as a subject consistency method.) and face-consistency methods, as well as several strong commercial open-/closed-source image generation systems.
As shown in~\cref{tab:quantitative_comparison}, existing controllable identity-preserving methods exhibit a clear trade-off between overall appearance consistency and facial identity fidelity. Subject-oriented methods (e.g., InstantCharacter~\cite{tao2025instantcharacter}) achieve relatively strong ReID performance but underperform in facial similarity, whereas face-centric methods (e.g., InfiniteYou~\cite{jiang2025infiniteyou}) preserve facial identity more effectively while sacrificing broader appearance consistency. A similar trade-off can also be observed in several commercial models. Among the compared commercial systems, close source method GPT-Image-2~\cite{gpt_image_2} delivers the strongest overall identity preservation, while the remaining models still lag behind our method in holistic identity consistency, especially facial fidelity.
In contrast, our method mitigates the instability of appearance consistency caused by facial dominance within naive dual-branch framework. This results in a more balanced trade-off between facial fidelity and appearance consistency, outperforming existing open-source model while remaining competitive with commercial systems, and providing an interpretable and trainable framework for holistic identity-preserving person image generation.

\begin{figure}[t]
\begin{center}
  \includegraphics[width=0.99\linewidth]{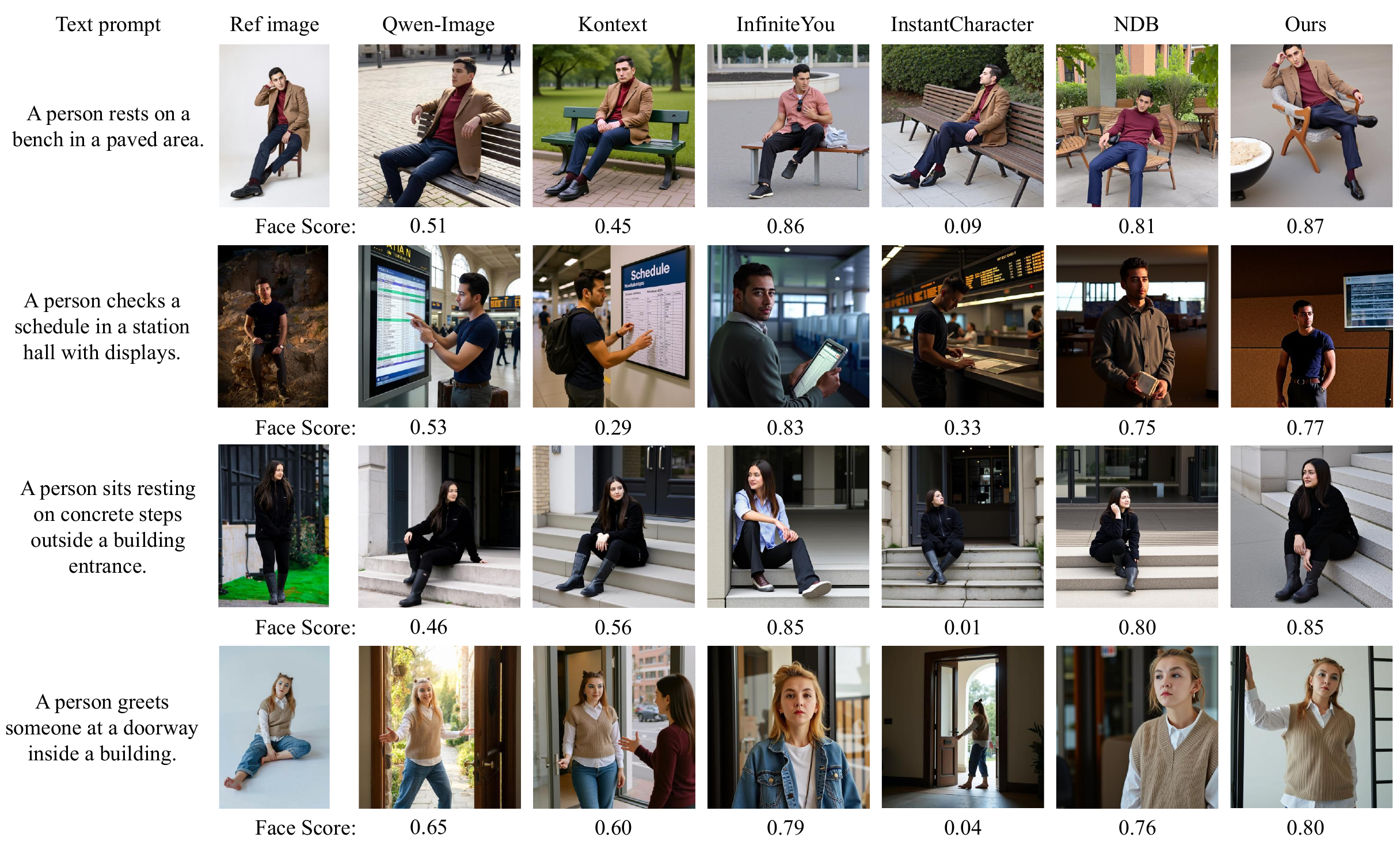}
\end{center}
  \caption{Qualitative comparison across different methods. ``NDB'' stands for Naive Dual-Branch. ``Face score'' represents the similarity between the generated image and the reference image. Our method achieves the best results in terms of holistic consistency (including appearance and face).  Please zoom in for better visualization. Best viewed in color.}
\label{fig:qualitative_comparison}
\end{figure}

\textbf{Qualitative comparison.}
We provide qualitative comparisons with representative open-source methods in~\cref{fig:qualitative_comparison} to visually assess identity consistency across facial and appearance cues.
As shown in the figure, different methods exhibit different trade-offs between facial similarity and overall appearance preservation. InfiniteYou~\cite{jiang2025infiniteyou} preserves facial identity slightly better than our method in some cases, but often fails to maintain consistent clothing and broader appearance details. In contrast, several other methods preserve overall appearance more reasonably, yet their generated faces are noticeably less similar to the reference identity. The naive dual-branch (NDB) partially improves this trade-off and generally maintains facial identity well, but still suffers from unstable branch coordination, with clothing details occasionally degrading or becoming inconsistent.
Compared with these methods, our approach achieves a more favorable trade-off between facial fidelity and appearance preservation. While not always yielding the strongest facial similarity, it better maintains both facial identity and overall appearance consistency simultaneously, resulting in more coherent identity preservation across diverse scenarios.

\subsection{Ablation Studies}
\textbf{Effectiveness of Component.}
We conduct component ablations to evaluate the contributions of Adaptive Temporal Gating (ATG) and Region-Aware Optimization (RAO) on top of the naive dual-branch baseline (NDB). As shown in~\cref{tab:Effectiveness_Component}, both ATG and RAO contribute to performance improvements to varying degrees, validating the effectiveness of the proposed components.
Notably, the full model does not achieve the highest Face and CLIP-T scores, which is in line with our expected goals. Specifically, the proposed DBS (ATG + RAO) is not designed to maximize facial similarity alone, but rather to improve holistic identity consistency by coordinating facial and appearance cues.
The relatively small variation in CLIP-T scores is also expected, as our method adjusts feature injection through lightweight gating, without modifying the backbone representation learning itself.
Overall, these results demonstrate that ATG and RAO are complementary, and their combination leads to more comprehensive identity preservation.

\begin{table}[t]
\centering
\begin{minipage}[t]{0.48\linewidth}
\centering
\caption{Effectiveness of components.}
\resizebox{\linewidth}{!}{%
\begin{tabular}{p{2.2cm}cccc}
\toprule
\textbf{Method} & \textbf{Face} & \textbf{ReID} & \textbf{DINOv2} & \textbf{CLIP-T} \\
\midrule
NDB        & 0.7746 & 0.9100 & 0.3024 & 0.1987 \\
NDB + ATG  & 0.7933 & 0.9113 & 0.3143 & 0.2017 \\
NDB + RAO  & 0.7808 & 0.9099 & 0.3031 & 0.2001 \\
\midrule
Ours       & 0.7875 & 0.9122 & 0.3202 & 0.2008 \\
\bottomrule
\end{tabular}%
}
\label{tab:Effectiveness_Component}
\end{minipage}
\hfill
\begin{minipage}[t]{0.48\linewidth}
\centering
\caption{Comparison of optimization strategies.}
\resizebox{\linewidth}{!}{%
\begin{tabular}{p{2.2cm}cccc}
\toprule
\textbf{Method} & \textbf{Face} & \textbf{ReID} & \textbf{DINOv2} & \textbf{CLIP-T} \\
\midrule
MSE      & 0.7801 & 0.9096 & 0.2974 & 0.1988 \\
LS       & 0.7638 & 0.9097 & 0.3080 & 0.2015 \\
Fairgrad & 0.7778 & 0.9106 & 0.3157 & 0.2003 \\
\midrule
Ours     & 0.7875 & 0.9122 & 0.3202 & 0.2008 \\
\bottomrule
\end{tabular}%
}
\label{tab:Optimization_Strategy}
\end{minipage}
\end{table}


\textbf{Analysis of Optimization Strategy.}
To examine the effect of the proposed region-aware optimization, we compare it with several alternative objective formulations, including: (1) MSE, which directly uses the original regional MSE objective; (2) LS, which linearly sums the face, appearance, and global losses; and (3) FairGrad~\cite{wang2025fairhuman}, which automatically adjusts regional loss weights based on gradient balancing.
As shown in~\cref{tab:Optimization_Strategy}, the compared alternatives exhibit different limitations in coordinating heterogeneous regional supervision. Direct MSE optimization lacks explicit handling of the distinct roles and scales of facial, appearance, and global regions. LS treats all objectives uniformly, which may lead to suboptimal coordination when these objectives compete during training. FairGrad introduces adaptive weighting, but it remains agnostic to the distinct semantic priorities of facial, appearance, and global supervision under cross-granularity interference. In comparison, our region-aware optimization provides a more consistent trade-off across regional objectives by maintaining global reconstruction


\section{Conclusion}
This paper addresses the problem of holistic identity consistency in personalized person image generation and introduces a naive dual-branch method (NDB) that integrates a subject (appearance) consistency branch and a face consistency branch. However, this simple combination suffers from generation instability due to heterogeneous inter-branch interactions and imbalanced signal strengths.
To address this issue, we further propose Dynamic Balancing Scaling (DBS), a fine-tuning strategy that improves face and appearance identity coordination through adaptive temporal gating and region-aware optimization. DBS achieves a better balance between facial fidelity and appearance consistency by mitigating the dominance of the facial branch and promoting more effective region-aware guidance.
We also introduce Pexels-100, a benchmark for evaluating holistic identity consistency. Extensive evaluation on this benchmark demonstrate that our method produces more coherent holistic identity preservation compared to existing open-source methods. It is worth mentioning that our method provides a controllable basic framework for holistic identity modeling.

\section{Limitations and Societal Impact}
\begin{figure}[ht]
\begin{center}
  \includegraphics[width=0.95\linewidth]{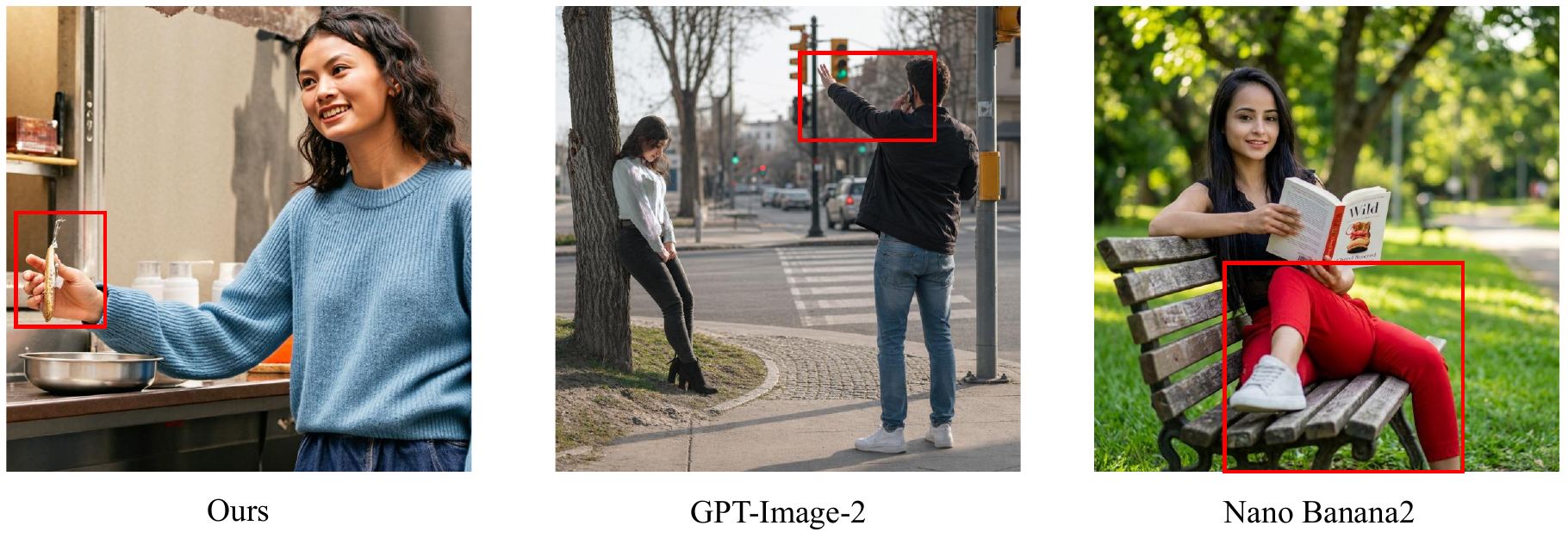}
\end{center}
  \caption{Modeling complex human poses or fine-grained structures remains challenging.}
  \label{fig:failurecases}
\end{figure}

\textbf{Limitations}. While the proposed method improves holistic identity consistency, it relies primarily on gating-based modulation and does not modify the underlying feature learning of the backbone model. As a result, it may still struggle in scenarios involving complex human poses or fine-grained structures, such as hands, where distortions or artifacts can occur. This limitation is largely inherited from the capacity of the underlying diffusion model and remains a common challenge in human image generation, as shown in~\cref{fig:failurecases}.
Future work may explore integrating stronger structural priors or geometry-aware refinement modules to enhance feature learning and further improve the modeling of complex structures.

\textbf{Societal Impact}. The ability to generate identity-consistent portraits enables applications in digital content creation and virtual media. However, it also raises concerns regarding privacy, portrait rights, and potential misuse for identity impersonation or deepfake generation. These risks may be mitigated by complementary efforts in AI-generated content detection and authentication, such as forgery detection and watermarking techniques. We advocate for the responsible use of such technologies within appropriate ethical and legal frameworks.

\bibliographystyle{plain}
\bibliography{ImageGEN}

\end{document}